\documentclass[10pt,journal,compsoc]{IEEEtran}
\usepackage[nocompress]{cite}
\usepackage{url}
\usepackage{amsmath,amssymb,graphicx}
\usepackage{lipsum} % Only used to generate random text.
\usepackage{algorithmicx}
\usepackage[ruled]{algorithm}
\usepackage{algpseudocode}
\usepackage[switch]{lineno}
\usepackage{multirow}
\usepackage[normalem]{ulem}
\usepackage{microtype}
\newcommand{\red}[1]{{\color{red} #1}}

\newcommand{\comment}[1]{}

\usepackage{colortbl}
\usepackage{relsize}
\usepackage{epsfig}
\usepackage{subcaption}
\usepackage{fixltx2e}
\makeatletter
\usepackage{pgfplots,pgfplotstable}
\usetikzlibrary{pgfplots.groupplots}
\usepackage{tikzscale}

\makeatletter
\DeclareRobustCommand\onedot{\futurelet\@let@token\@onedot}
\def\@onedot{\ifx\@let@token.\else.\null\fi\xspace}

\makeatother

\newcommand*\rot{\rotatebox{90}}

\newcommand{\boldparagraph}[1]{\vspace{0.2cm}\noindent{\bf #1:} }

\usepackage{xcolor}
\usepackage{hyperref}
\usepackage{pifont}% http://ctan.org/pkg/pifont
\newcommand{\cmark}{\ding{51}}%
\newcommand{\xmark}{\ding{55}}%
\ifCLASSINFOpdf
\else
\fi
\begin{document}
\title{Light Field Implicit Representation for \\ Flexible Resolution Reconstruction}

\author{Paramanand Chandramouli, Hendrik Sommerhoff, Andreas Kolb
\IEEEcompsocitemizethanks{\IEEEcompsocthanksitem The authors are with with the Department
of Computer Science, University of Siegen, Siegen 57076.}\protect\\
% note need leading \protect in front of \\ to get a newline within \thanks  .
{\small E-mail: {\{paramanand.chandramouli, hendrik.sommerhoff, andreas.kolb\}@uni-siegen.de}
}}
\markboth{Journal of \LaTeX\ Class Files,~Vol.~14, No.~8, August~2015}%
{Shell \MakeLowercase{\textit{et al.}}: Bare Demo of IEEEtran.cls for Computer Society Journals}
\IEEEtitleabstractindextext{%
\begin{abstract}
\comment{Inspired by the recent advances in implicitly representing signals with trained neural networks, we aim to learn a continuous representation for narrow-baseline 4D light fields. We propose an implicit representation model for 4D light fields which is conditioned on a sparse set of input views. Our model is trained to output the light field values for a continuous range of query spatio-angular coordinates. Given a sparse set of input views, our scheme can super-resolve the input in both spatial and angular domains by flexible factors. consists of a feature extractor and a decoder which are trained on a dataset of light field patches. The feature extractor captures per-pixel features from the input views. These features can be resized to a desired spatial resolution and fed to the decoder along with the query coordinates. This formulation enables us to reconstruct light field views at any desired spatial and angular resolution. Additionally, our network can handle scenarios in which input views are either of low-resolution or with missing pixels. Experiments show that our method achieves state-of-the-art performance for the task of view synthesis while being computationally fast.% Our scheme is robust and can handle varying levels of spatial sparsity with a single model even in the absence of correspondence across input views. Using only about 2\% of the total number of pixels of a light field sample, we can synthesize views of reasonably good quality.
}
Inspired by the recent advances in implicitly representing signals with trained neural networks, we aim to learn a continuous representation for narrow-baseline 4D light fields. We propose a novel  implicit model for 4D light fields  conditioned on convolutional features of a sparse set of input views. This conditioning enables our model to generalize across scenes without retraining the implicit model for each scene. Our model can be queried at continuous 4D light field coordinates, allowing joint spatial-angular super-resolution at flexible super-resolution factors.   We demonstrate the flexibility of the proposed method  with experiments on  the tasks of view synthesis,  joint spatial-angular super-resolution on real light fields. Our  model outperforms current state-of-the-art baselines on these tasks, while utilizing only a fraction of run-time of the baselines.  Further, our model can also be trained to be robust to varying levels of missing pixels in the input views.
\end{abstract}

%\begin{IEEEkeywords}
%Computer Society, IEEE, IEEEtran, journal, \LaTeX, paper, template.
%\end{IEEEkeywords}
}

\maketitle

\IEEEdisplaynontitleabstractindextext

\IEEEpeerreviewmaketitle

\IEEEraisesectionheading{\section{Introduction}\label{sec:introduction}}
\begin{figure*}[htb]
\begin{center}
%\hspace{-10pt}
\footnotesize
\resizebox{\linewidth}{!}{
\begin{tabular}{cccccc}
\includegraphics[width=90pt]{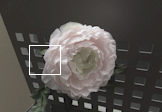}&\hspace{-10pt}
\includegraphics[width=85pt]{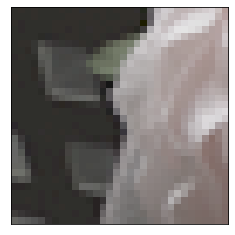}&\hspace{-10pt}
\includegraphics[width=85pt]{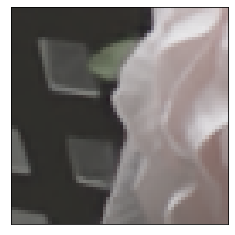}&\hspace{-10pt}
\includegraphics[width=85pt]{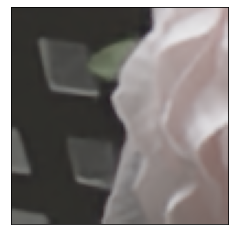}&\hspace{-10pt}
\includegraphics[width=85pt]{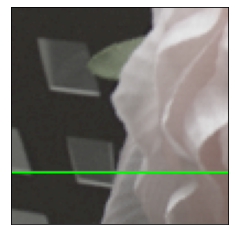}&\hspace{-10pt}
\includegraphics[width=85pt]{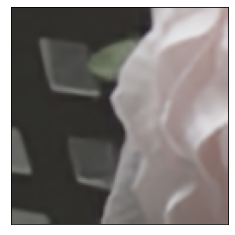}\\

(a) Low-res(LR) view & \hspace{-10pt} (b) LR Patch & \hspace{-10pt} (c) ($\times 2$ spatial) & \hspace{-10pt} (d) ($\times 3.3$ spatial) & \hspace{-10pt} (e) Ground-truth patch & \hspace{-10pt} (f) ($\times 4$ spatial)
\end{tabular}}

\begin{tabular}{cccc}
\includegraphics[width=110pt]{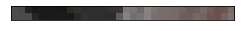}&\hspace{-0pt}
\includegraphics[width=110pt ]{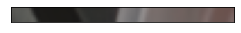}&\hspace{-0pt}
\includegraphics[width=110pt]{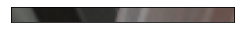}&\hspace{-0pt}
\includegraphics[width=110pt]{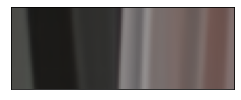}\\
(g) LR input EPI & (h) ($\times 3.5$ angular) & (i) Ground-truth EPI & (j)($\times 19.5$ angular)
\end{tabular}
\caption{Flexible resolution reconstructions from our single Conditional Implicit Light Field Network (CILN). The input to our model is a set of $4$ corner views selected from the angular grid of size $7{\times}7$. The input views were downsampled even in spatial domain by a scale factor of $3.3$. Our \emph{single} model produces high quality reconstructions at different super-resolution factors in both spatial and angular domain. In the bottom row, we illustrate the epipolar plane image (EPI) obtained at the vertical location indicated by the green line in Fig.\ref{fig:teaser} (e). While Figs. \ref{fig:teaser} (c), (d) and (f) show flexible spatial domain reconstruction, the EPI images in Figs. \ref{fig:teaser} (h) and (j) show angular super-resolution. 
 }\label{fig:teaser}
 \vspace{-5pt}
\end{center}
\end{figure*}
% Pixel aligned feature representations are extracted from sparse 4D light field input views using a CNN. Our network outputs intensity values (RGB) corresponding to the queried angular location $(s,t)$ at each pixel location $(x,y)$ by utilizing the conditioned feature embedding $\phi(\mathbf{o},(x,y))$

%such as free viewpoint rendering of a 3D scene, with applications in

%Cheaper alternatives to small base-line light field acquisition project light fields onto the 2D image sensor using lens arrays, this however has a trade-off between spatial and angular resolution of the captured light field.
\IEEEPARstart{A}{cquisition} and representation of high quality light fields is important in many diverse applications such as virtual reality, microscopy and computational photography.  Light fields~(LFs) are characterized using a continuous four dimensional function which represents the scene radiance along spatial and angular coordinates. 
    As light fields dimensionally scale as $O(n^4)$, acquiring  and storing densely-sampled LFs is expensive and  challenging. Hence, several approaches have been developed to computationally reconstruct dense LFs from sparse samples \cite{wanner2013variational,9106041}. Recently, several deep learning-based methods have been developed for efficient reconstruction of photo-realistic novel views from sparse views \cite{kalantari2016learning,wuTIP2019}. These approaches are often trained for a specific configuration of input observations and output resolution and offer limited flexibility.%However, these approaches often offer limited flexibility in terms of input view locations, and they provide novel views which are limited to a fixed  grid, and typically require retraining to achieve a denser sampling.

An emerging class of neural signal representation methods referred to as implicit representations have attracted  high research interest as of late \cite{sitzmann2019siren}. These techniques provide a continuous function model for  signals by  parameterizing them through multi-layer perceptrons (MLPs) which are composed of deep fully connected neural networks. Apart from leading to an efficient representation, implicit representations offer the flexibility of rendering the signal value at any desired input location. Recent works, including \cite{sitzmann2019siren,tancik2020fourfeat} have demonstrated remarkable abilities of trained neural networks in implicitly representing different classes of signals such as images, videos and 3D shapes. However, these works typically require training a separate network to represent  each scene.%\cite{Bemana2020xfields} propose to learn implicit representation of a scene across of time, view, and lighting dimensions with limited observations by including a differentiable rendering step into the network architecture.
%The methods by ,\cite{mildenhall2020nerf} have demonstrated excellent results in rendering complex scenes with high fidelity. Time varying lighting and views \cite{Bemana2020xfields}

A few techniques have been developed in recent literature  to generalize the implicit neural representations to a class of signals, without retraining from scratch for each signal.   Examples of such methods include the use of meta-learning \cite{NEURIPS2020_731c83db,tancik2020learned}  or training a hyper-network \cite{sitzmann2019scene,sitzmann2019siren}  for initial network weight generation. These methods however require further test-time optimization of weights for each signal.  An alternate approach  to avoid retraining  is to train the implicit network by providing features from the representation space in addition to input coordinates in a supervised fashion. Such an approach has been explored in context of 3D reconstruction and shape representation in \cite{mescheder2019occupancy, park2019deepsdf, NEURIPS2019_disn, Peng2020ECCV} for image super-resolution \cite{chen2020learning} and scene representation~\cite{yu2020pixelnerf}.

 In this work, we propose to learn a conditional implicit representation for 4D light fields which simultaneously can render scene radiance at continuous 4D query coordinates, while generalizing across different scenes, without the need of retraining. %We consider that a very small set of light field views are available as inputs to our model.
 We develop a conditional implicit light field network (CILN) which is conditioned on very sparse set of input views to generate such implicit light field representation. Our formulation is  similar to the approaches of~\cite{chen2020learning,NEURIPS2019_disn,Peng2020ECCV}, applied to LF data. Our model consists of a convolutional feature extractor and an implicit decoder. %Fig.~1~(a) provides an overview of our approach. 
 The CNN feature extractor embeds the input spatio-angular contextual information into the representation space. For flexibly decoding at any desired spatial resolution, the extracted features are resized to the desired resolution leading to a per-pixel latent feature representation. The 4D  scene radiance is provided by an MLP decoder that is dependent on both the per-pixel features and the query spatio-angular coordinates. The use of a CNN feature extractor together with a conditioned decoder allows for scalable and robust implicit representations facilitating the reconstruction of fine-grained detail. Our approach allows  super-resolution of input views simultaneously in both spatial and angular domains by flexible super-resolution factors. %Additionally, our model can also be trained to handle scenarios wherein the input views  have spatially missing pixels. 

  %Both the CNN feature extractor and the MLP decoder are jointly trained with a simple maximum likelihood learning of the parameters.

In Fig.~\ref{fig:teaser}, we show a sample result of LF view reconstruction using our approach. The inputs to our model are the four corner views that are also of low spatial resolution. Our CILN model is able to reconstruct good quality LF views at any desired spatial and angular resolution. In Figs. \ref{fig:teaser} (c), (d) and (f), we show the output of our model at different spatial resolutions. To illustrate angular super-resolution, we show EPIs \cite{wanner2013variational} in the bottom row of Fig. \ref{fig:teaser}. These EPIs show the LF plotted against horizontal spatial and horizontal angular coordinates for a specific fixed vertical location. Our reconstructions for different angular resolutions are shown in Figs. \ref{fig:teaser} (h) and (j). 

\comment{Our proposed model can  synthesize high quality LF views outperforming the state of the art methods for LF view synthesis, and joint spatial-angular super-resolution. Our model can be trained to handle different input view configurations, with as low as only two input views. Further, our model generalizes well to infer angular resolutions not used in training.  Using a simple architecture, our model synthesizes high quality views,  while utilizing only a fraction of run-time of competing view synthesis methods. In addition, our model can also be trained to handle challenging scenarios where both the spatial and angular measurements are sparse, with varying levels of missing pixels, which cannot be handled well by competing view reconstruction methods.}%\\
Our main contributions are summarized below:
\begin{itemize}
\item We propose a novel conditional implicit representation for 4D light fields which can be queried at continuous coordinates, allowing \emph{joint spatial-angular super-resolution} at \emph{flexible} super-resolution factors.
%\item We propose a conditional implicit representation for4D light fields which enables light field synthesis at continuous spatio-angular coordinates.
\item Our approach generalizes across scenes without requiring retraining.
%\item  Our model can be trained to handle joint spatial and angular super-resolution in which input views can be flexibly super-resolved in spatio-angular domain.
\item Our model achieves state of the art results on small baseline view synthesis and spatial angular super-resolution.
\item Using a simple architecture, our model synthesizes high quality LF views, while utilizing only a fraction of run-time of competing view synthesis methods.
\item Our model can be trained to handle challenging scenarios, where both the spatial and angular measurements are sparse, with varying levels of missing pixels.%, which cannot be handled  by competing methods. 
\end{itemize}
%4) Can render views at a higher spatial resolution than the input at test time without the requiring additional training.
\comment{\red{The main contributions are summarized below:
\begin{itemize}
\item We propose a conditional implicit representation for LFs which enables  light field reconstruction on a continuous domain of spatio-angular coordinates.
\item The proposed approach generalizes across scenes and does not require retraining.
\item Our model can be trained to handle joint spatial and angular super-resolution in which input views can be flexibly super-resolved in spatio-angular domain.
\item Our simple architecture leads to low computation times and yet performs quite well.

\end{itemize}}}
\section{Related work}
\textbf{LF View Interpolation and Synthesis:~}
Recovery of dense LFs from sparse views is highly challenging. Many techniques have been proposed to tackle the problem of LF view interpolation also referred to as angular super-resolution. %., LF view synthesis or LF reconstruction. 
A good overview of existing approaches for view synthesis is available in \cite{9106041}. We note that techniques have been proposed to address view synthesis for large baseline LFs such as \cite{mildenhall2019local}. However, we restrict our discussion and comparison to the works which consider small baseline LFs. Traditional methods for view interpolation exploit light field geometry information \cite{wanner2013variational} or sparsity of LFs, for e.g in Fourier domain \cite{shi2014light},    in learned dictionary \cite{schedl2015directional,schedl2018optimized} or in sheared-EPI representation \cite{vagharshakyan2017light}  for variational LF reconstruction.  Starting from \cite{kalantari2016learning}, several deep learning based solutions \cite{GenModelPAMI,wang2018end, wing2018fast, meng2019high, jin2019flexible, wuTIP2019, WuEPICNN2019, jin2020learning, Shi_2020_CVPR, wu2020spatial} have been proposed for synthesizing dense LFs. While some of these methods \cite{meng2019high,wing2018fast, wang2018end} employ CNNs to directly regress dense LFs from input views, others incorporate additional geometric information such as EPI structure \cite{wuTIP2019,WuEPICNN2019}  or disparity-based warping \cite{  kalantari2016learning,jin2019flexible,jin2020learning,Shi_2020_CVPR} into their network architecture. While EPI-based methods \cite{wuTIP2019,WuEPICNN2019} require input views on a regular grid, warping based methods can operate on irregular input views, but they typically  need to be trained for each input configuration and output resolution separately.  Recently, techniques \cite{GenModelPAMI,jin2019flexible} have been proposed that can generate dense LF from a flexible set of input views. While \cite{GenModelPAMI}  proposes to optimize the latent code of a LF generative model to  fit the inputs and observation model,  \cite{jin2019flexible} use a plane sweep volume for disparity estimation from a flexible pattern of fixed number of input views.  However, all the approaches \cite{kalantari2016learning, wang2018end, wing2018fast, meng2019high, jin2019flexible, wuTIP2019, jin2020learning, Shi_2020_CVPR, wu2020spatial} explicitly or implicitly assume the existence of correspondence across different input views and cannot handle spatial sparsity with exception to \cite{GenModelPAMI} which requires expensive optimization for view interpolation, and can only generate views of fixed angular resolution. %While the approach  of  \cite{jin2019flexible} can generate flexible output views, 
Our approach can generate views of flexible spatial and angular resolution and can also be trained to handle spatial sparsity.\vspace{2pt} \newline
\textbf{LF Spatial Super-resolution:~}Many approaches have been developed to overcome limited spatial resolution in LFs, some recent works include~\cite{Jin_2020_CVPR,9286855}. While these works focus on achieving super-resolution in spatial domain only, we address a more challenging task of joint spatio-angular super-resolution by flexible factors. We note that joint spatio-angular super resolution has been attempted in the work of Meng~et~al. \cite{meng2019high}, for small and fixed spatial and angular super resolution factors.\vspace{2pt}\\
\begin{figure*}[htb]
\begin{center}
\includegraphics[width=\textwidth]{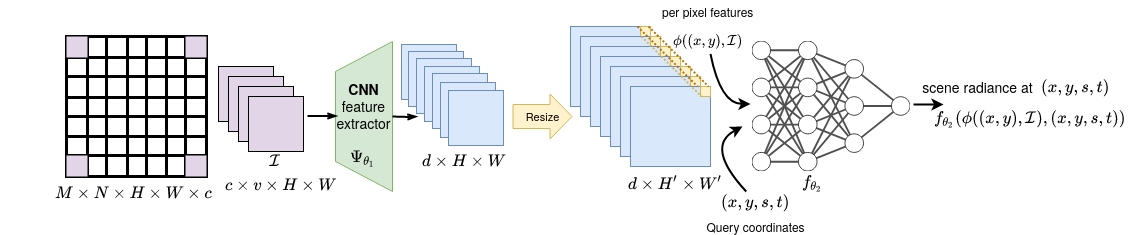} 
\caption{Overview of  Conditional Implicit Light Field Network}\label{fig:CILN}
\vspace{-5pt}
\end{center}
\end{figure*}
\textbf{Implicit Representations:~}
Recent research \cite{sitzmann2019siren, tancik2020fourfeat} has demonstrated that implicit parameterizations of continuous functions using trained multi layer perceptrons as a powerful and efficient alternative to conventional representations. %Recent works have demonstrated impressive abilities of trained MLPs to achieve an accurate implicit signal representation. 
%\red{These works \cite{sitzmann2019siren, tancik2020fourfeat} employ an additional Fourier encoding of coordinates, or with sinusoidal activation functions to enable the implicit network to capture fine grained details.}
Applications of implicit representations  have been shown for modeling shapes and objects \cite{genova2019learning, gropp2020implicit}, or scenes~\cite{sitzmann2019scene, Bemana2020xfields, mildenhall2020nerf, sitzmann2019siren},  time-varying 3D geometries~\cite{Niemeyer2019ICCV} and video and waveform representation \cite{sitzmann2019siren}, and more recently for 4D LFs in \cite{Feng21} using a novel input coordinate transformation strategy. Some of these works, including ~\cite{sitzmann2019scene, Niemeyer2019ICCV, mildenhall2020nerf, Bemana2020xfields} make use of differentiable rendering \cite{tewari2020state}  to learn complex scene geometries using a sparse set of input views.  Additionally, \cite{ Bemana2020xfields} propose to capture variations across view, time and lighting conditions by learning compact implicit neural representations with a learnable view-time geometry model with hard-coded  differentiable rendering and warping.  While all these methods have demonstrated excellent results in representing a  signal, object or scene, they need retraining for each new instance to achieve faithful representation.\vspace{2pt}\\
\textbf{Generalizing Implicit Representations:~}
To avoid retraining the implicit network from scratch for each new instance  \cite{NEURIPS2020_731c83db,tancik2020learned}  use  meta learning  to learn initial network weights for a class of shapes or signals to achieve relatively faster inference times. However, network weights need further optimization at test time for each new instance for high quality representation. An alternate approach to generalization is representing instances of signals as latent codes.  Hypernetworks~\cite{ha2016hypernetworks} and conditional implicit networks for 3D shapes~\cite{mescheder2019occupancy, park2019deepsdf}  use a single low dimensional latent code for each instance of signal. Closely related to such approaches are the conditional neural processes~\cite{garnelo2018conditional}.   Given a set of observations and the corresponding locations, referred to as context-set, the  conditional neural processes~\cite{garnelo2018conditional} and  hypernetworks~\cite{ha2016hypernetworks, sitzmann2019scene,sitzmann2019siren}  aggregate latent embeddings extracted by an encoder of all inputs, which are provided to a hypernetwork to generate implicit network weights \cite{sitzmann2019scene} %a hypernetwork is trained to map such aggregated latent embedding to implicit network weights, which can be further optimized using few shot reconstruction.
or used to condition an MLP decoder by concatenating with the queried point as input%Conditional neural processes~
\cite{garnelo2018conditional}.
%instead condition the MLP decoder by providing as input the latent embedding together with the query point. 
Similar input concatenated latent code conditioning is exploited by the implicit networks of~\cite{mescheder2019occupancy, park2019deepsdf}.  While the approach of ~\cite{mescheder2019occupancy} utilizes an encoder to directly map the entire context-set to a latent embedding, the auto-decoder approach of ~\cite{park2019deepsdf}  does not have an encoder and requires optimization at test time to find the optimal latent embedding corresponding to the input observations.  

Though use of a single latent code  for representing each instance in a class of signals, leads to a compact latent representation,  such an approach leads to underfitting the context information, and therefore cannot capture fine details. Alternatively,  recent approaches \cite{saito2019pifu, NEURIPS2019_disn, Chibane_2020_CVPR, Peng2020ECCV}  utilize CNNs to generate a tensor of  feature embeddings which are functions of both the input coordinates and observations.  The implicit network then decodes the input query point using the corresponding feature embedding, which allows representation of higher resolution details. While \cite{saito2019pifu, NEURIPS2019_disn,chen2020learning,yu2020pixelnerf} operate on image data and therefore use pixel aligned implicit representations, \cite{Chibane_2020_CVPR, Peng2020ECCV} can operate on  the 3D point clouds, whose encodings are discretized to regular grid for further processing by CNNs. 

While most  conditional implicit networks have focussed on 3D shape representation and reconstruction, % for e.g. using hypernetworks \cite{sitzmann2019scene}, using latent code conditioned implicit networks  \cite{park2019deepsdf, NEURIPS2020_731c83db, NEURIPS2019_disn, mescheder2019occupancy, Peng2020ECCV}, and for reconstruction textured 3D surfaces of clothed humans \cite{saito2019pifu}, and 3D object appearance~\cite{Oechsle2020THREEDV}. 
 recent works  \cite{chen2020learning,yu2020pixelnerf} have developed  such representations for arbitrary super-resolution of images and scene representation respectively. %\red{discuss PixelNerf \cite{}} .\red{ We extend their approach for 4D light felds. Flexible super-resolution.}
Since LF views are available on a regular grid, we utilize pixel aligned representations similar to \cite{saito2019pifu, NEURIPS2019_disn,chen2020learning,yu2020pixelnerf}.  %While \cite{saito2019pifu, NEURIPS2019_disn} learn implicit networks to predicts occupancy and signed distance fields respectively for 3D reconstruction, we learn an implicit function to predict scene radiance.
While \cite{saito2019pifu, NEURIPS2019_disn,chen2020learning} learn the feature extractor, \cite{yu2020pixelnerf} uses imagenet pretrained network features. 
 %The CNN encoder in \cite{NEURIPS2019_disn}  obtains a concise a global latent code for all points in the input. Further, local features corresponding to pixels are extracted from each layer of the encoder to aid  fine grained reconstruction.  
To achieve flexible spatial resolutions, we spatially upsample features similar to \cite{chen2020learning}. To capture the context across views, we learn a fused embedding at pixel level by training our feature extractor using concatenated input views, while  \cite{saito2019pifu,yu2020pixelnerf} average the features from each view point  to obtain an aggregated embedding.\\
\textbf{Recovery from Missing Pixels:~}
Image recovery  from  sparse  pixels  has  been studied since several years e.g \cite{liu2012tensor,li2016marlow,yeh2016semantic}. Physically random  pixel  sampling  is  supported in CMOS based image sensors \cite{scheffer1997random}  found in hand held cameras.  While traditional methods to restore missing pixels \cite{liu2012tensor,li2016marlow} can operate on varying levels of sparsity, deep networks  often  cannot handle varying levels of corruption with a single network~\cite{gao2017demand}. Deep learning based  image recovery from varying levels of missing pixels has been demonstrated in ~\cite{garnelo2018conditional} for a class of images, e.g. face images. However, the reconstructions are blurry due to underfitting.
%\red{Recent work demonstrates that inpainting methods as an alternative to traditional image compression, where a tiny fraction of well chosen pixels can be used to reconstruct the image using interpolation.}
 Similar results have not been shown for LF recovery.
\section{Proposed Method}
%In this paper, we propose an implicit representation of 4D light fields through a conditional implicit light field network (CILN). 
Let 4D light fields be denoted by the continuous function  $L(x,y,s,t)$  representing the scene radiance at spatial coordinates  $(x,y)$ and angular coordinates $(s,t)$. Our aim is to approximate this function implicitly using a trained neural network, which generalizes across scenes. To achieve this generalization, we condition the implicit neural network on features of a small set of input views extracted from a CNN. Fig.~\ref{fig:CILN} provides an overview of the proposed \emph{conditional implicit light field network (CILN)} framework, which consist of two parts: \emph{feature extractor} $\Psi_{\theta_1}$ and \emph{scene radiance decoder $f_{\theta_2}$}. 
\begin{figure*}[h]
 \begin{center}
 \resizebox{\linewidth}{!}{
\begin{tabular}{ll ll ll ll lll}
\multicolumn{2}{c}{\includegraphics[width=150pt]{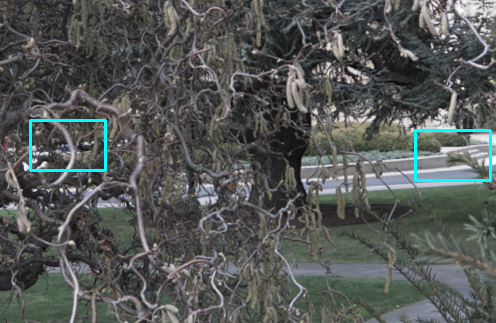}}&
\multicolumn{2}{c}{\hspace{-12pt}\includegraphics[width=150pt]{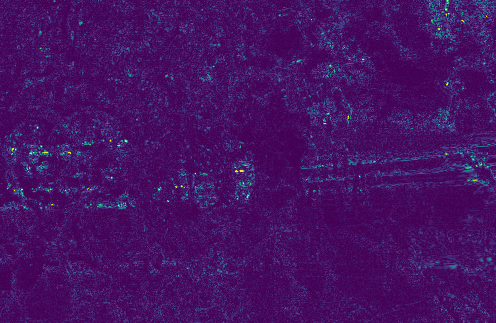}}&
\multicolumn{2}{c}{\hspace{-12pt}\includegraphics[width=150pt]{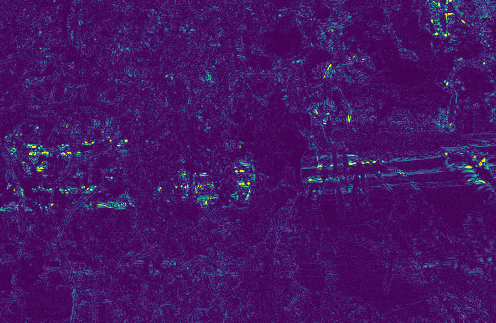}}&
\multicolumn{2}{c}{\hspace{-12pt}\includegraphics[width=150pt]{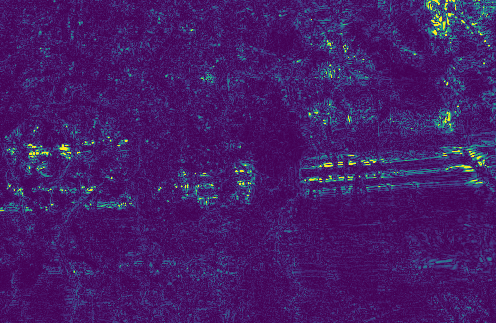}}&
\multicolumn{2}{c}{\hspace{-12pt}\includegraphics[width=150pt]{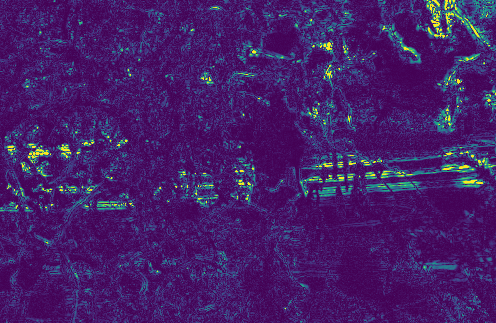}}&
\hspace{-12pt}\includegraphics[width=25pt,height=105pt]{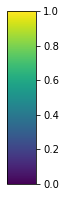}\tabularnewline
\includegraphics[width=74pt]{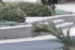}&\hspace{-12pt}
\includegraphics[width=74pt]{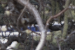}&\hspace{-12pt}
\includegraphics[width=74pt]{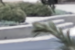}&\hspace{-12pt}
\includegraphics[width=74pt]{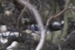}&\hspace{-12pt}
~\includegraphics[width=74pt]{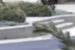}&\hspace{-12pt}
\includegraphics[width=74pt]{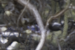}&\hspace{-12pt}
\includegraphics[width=74pt]{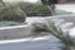}&\hspace{-12pt}
\includegraphics[width=74pt]{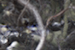}&\hspace{-12pt}
\includegraphics[width=74pt]{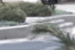}&\hspace{-12pt}
\includegraphics[width=74pt]{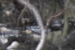}&\hspace{-12pt}\vspace{2pt}\\
\multicolumn{2}{c}{\large Ground truth} &\multicolumn{2}{c}{\large Ours} 
&\multicolumn{2}{c}{\large  Jin~et~al.\cite{jin2019flexible}}
&\multicolumn{2}{c}{\large  Meng~et~al.\cite{meng2019high}} 
&\multicolumn{2}{c}{\large Wu~et~al.\cite{wuTIP2019}}
\end{tabular}}
\caption{Comparison of synthesized central view of the LF `Occlusions16', for $2\times2\to7\times7$ view upsampling. First row shows the comparison of CILN reconstruction with the ground truth and recovered views using \cite{jin2019flexible,meng2019high,wuTIP2019}.  Second row shows zoomed-in patches corresponding to the marked regions. Error maps are depicted with error magnified by a factor of $5$.\vspace{-0.5pt}}
\label{Fig:2x2_7x7}
\end{center}
\end{figure*}
In practice, LF acquisition is done on a discrete grid, with fixed spatial and angular resolutions. We assume that the input $\mathcal{I}$ to the CNN feature extractor $\Psi_{\theta_1}$ are $v$ views having $c$ color channels sampled from an angular grid of size $M{\times}N$  and spatial extent $H\times W$. To obtain a feature representation, which simultaneously captures global context across views, we concatenate the input views along the color channel dimension, and train our feature extractor to learn a fused representation for the input views. %The input views denoted by $\bo$ are fed to a feature extractor.
We find that a simple CNN residual architecture with only two dimensional convolutions can efficiently provide feature representations. Further, the use of convolutions  allows us to handle light fields of varying spatial extent, and automatically endows  our output features with translational equivariance. Using appropriate padding in the convolutional layers, we obtain feature representation $\Psi_{\theta_1}(\mathcal{I})$ of dimension $H\times W\times d$, where $d$ is the per-pixel feature dimension. To facilitate reconstruction at flexible spatial resolutions, we resize the features to the desired resolution $H'{\times}W'$, the resized features are denoted by $\phi(\mathcal{I})$.

We utilize a multilayer perceptron (MLP) $f_{\theta_2}$ to implicitly parameterize the scene radiance function. Given a query point with spatio-angular coordinates $(x,y,s,t)$, the implicit decoder $f_\theta$ predicts the scene radiance (RGB values) conditioned on the per-pixel feature representation $\phi(\mathcal{I},(x,y))$, i.e the light field $L(x,y,s,t)$ can be represented as %$f_{\theta}\left( \phi(\mathcal{I},(x,y)), \left(x,y,s,t\right) \right)$.
\begin{align}
L\left(x,y,s,t\right) = f_{\theta_2}\left( \phi(\mathcal{I},(x,y)), \left(x,y,s,t\right) \right)
\end{align}
 
%Unlike many recent view synthesis methods which synthesize light fields on a fixed angular grid, our CILN can produce light fields with arbitrary angular resolution. 
%\newline
%\textbf{Conditional Representation for 3D Light Field} In addition to representation of 4D light fields, we also consider learning a conditional representation of light fields varying along a single angular dimension. We refer to this as  conditional implicit network for 3D light fields (CILN-3D). The input  to CILN-3D is a light field patch of dimension $3\times H\times W \times c$, indicated by purple patches in Fig.~\ref{fig:3D schematic}. The feature extraction and scene radiance prediction are similar to our original CILN formulation. The only difference is that the views are queried along a single angular dimension.
%EPI Loss
\boldparagraph{Implementation}
From a ground-truth LF patch of size $M{\times}N{\times}H'{\times}W'$, a fixed set of $v$ views are selected as input $\mathcal{I}$. %the input $\bo$ is obtained by selecting a fixed set of $V$ views from the $M{\times}N$ grid. 
 %In order to facilitate learning of flexible spatial super-resolution, we spatially downsample the $V$ views by different factors \cite{chen2020learning}.
To facilitate  flexible spatial super-resolution, we spatially downsample the input by different down sampling factors during training similar to \cite{chen2020learning}. Features are resized using bilinear interpolation.  We jointly train the feature extractor $\Psi_{\theta_1}$ and the implicit network $f_{\theta_2}$  to reconstruct light fields from the latent embedding, by minimizing loss between the ground truth pixel values and the reconstructions. The loss function  is a combination of L1 loss and  EPI gradient loss \cite{jin2020learning} which encourages
preservation of LF parallax structure.\\
%We train the feature extractor and the implicit network together to reconstruct a light fields from the latent embeddings, by minimizing a loss between the ground truth pixel values and the reconstructions.
%We train our model, on patches of light fields. The ground-truth patch $M{\times}N$.
%The ground truth label is the whole $7{\times}7$ grid of views. 
%The feature extractor embeds the context information from the input views into a latent feature representation of dimension 128, which is aligned with the spatial grid of input views.
%\textit{Architecture:~}
The feature extractor consists of four residual 2D CNN blocks \cite{he2016deep} followed by a 1D convolutional layer. The implicit decoder is an MLP consisting of 2 hidden layers of dimension 320. We will make our code and trained models publicly available subsequently.
% To train For obtaining ground-truth and input pairs with different

\section{Experiments}
    \begin{table}[b]
        \label{tab:2x2-7x7}
        \centering
        \begin{tabular}{c c c c }
        \hline
        Method & 30scenes & Occlusions & Reflective\\
        \hline
        Wu \textit{et al.}\cite{wuTIP2019}&39.17/0.975&34.41/0.955& 36.38/0.944\\
        Meng \textit{et al.}\cite{meng2019high}&40.18/0.975     &36.69/0.969      &37.59/ 0.952\\\
        Yeung \textit{et al.}\cite{wing2018fast}&42.77/\textbf{0.986}&38.88/0.980& 38.33/\textbf{0.960}\\
        Kalantari \textit{et al.}\cite{kalantari2016learning}&41.40/0.982&37.25/0.972&38.09/0.953\\
        Jin \textit{et al.}\cite{jin2019flexible}&42.75/\textbf{0.986}&38.51/0.979&38.35/0.957\\        Ours&\textbf{42.80/0.986}&\textbf{39.36/0.981}&\textbf{39.13/0.960} \\
        Ours\textsuperscript{$\dagger$}&41.50/0.983&38.44/0.978&38.61/0.958\\
        Ours\textsuperscript{$\dagger\dagger$}&42.34/0.985&38.83/0.979&38.89/0.960\\
        %2D CNN\\
        \hline
        \end{tabular}
        \caption{Quantitative comparisons (PSNR/SSIM) of  proposed CILN with the state-of-the-art view synthesis approaches for  $2\times2\rightarrow7\times7$ view interpolation.$\dagger$ indicates model trained for flexible spatial angular super-resolution. $\dagger\dagger$ indicates model trained for variable pixel sparsity.}
        \label{tab:2x2-7x7}
        \end{table}
We evaluate our approach on different LF recovery tasks: \\
\emph{i)~View interpolation:~} We train and evaluate our CILN model for view upsampling from $2\times2$ views to %task with different input configurations for 
recover $7\times7$ LF. Further, to demonstrate the flexibility of our approach, we apply the model trained for $2\times2\to7\times7$ LF recovery task for reconstructing $8\times8$ LFs without any retraining.\\
\emph{ii)~Spatial angular super-resolution:} We train CILN for flexible spatial resolutions for $7\times7$ LF recovery, this model is indicated as `Ours\textsuperscript{$\dagger$}' in  the experiments. During training, we downsample input LF patches by randomly chosen scale factors ranging between $0.25$ and $1$. We evaluate this model for flexible spatial and angular resolutions.\\
\emph{iii)~ LF recovery from sparse spatio-angular measurements:} We train and evaluate the CILN with varying levels of %pixel sparsity ranging from $0\%-90\%$ 
pixels missing from the $2\times2$ input views for $7\times7$ LF recovery. The extent of missing pixels in the input views is randomly chosen to lie in range $0\%-90\%$ during training. This model is indicated as `Ours\textsuperscript{$\dagger\dagger$}' in  the experiments.\\
We used the training set of Kalantari \emph{et al.}~\cite{kalantari2016learning}, consisting of real light fields captured using a Lytro camera. %, and the training set in the synthetic New HCI dataset~\cite{HCI_data}. CILN-3D is trained on the synthetic datasets~\cite{HCI_data,heber2016convolutional.
We evaluate our approach on the 30 scenes of Kalantari et al.'s test set, and the selected scenes (following \cite{jin2019flexible}) from the `reflective' and `occlusion' categories of the Stanford Lytro light field archive~\cite{Stanford_Lytro}.% The training settings of our models are provided in the supplementary material.
\subsection{View Interpolation}
We quantitatively validate the performance of our approach for LF recovery from sparse input views using average  PSNR and SSIM values of novel synthesized views 
and  provide visual comparisons of the view reconstructions using error maps with respect to ground truth.\vspace{1pt}\\
\textbf{{Fixed view interpolation $2\times2\to7\times7$:~}} Tab.~\ref{tab:2x2-7x7} provides a quantitative comparison between our method and the following baselines : i)~fully convolutional approaches of Meng et~al. \cite{meng2019high} and Yeung et~al. \cite{wing2018fast} ii)~warping based LF synthesis networks of Jin et~al. \cite{jin2019flexible} and Kalantari et~al. \cite{kalantari2016learning} iii)~deep network of Wu et~al. \cite{wuTIP2019} incorporating sheared EPI structures,  for the three datasets considered. All the baselines are trained and tested for the task of $2\times2\to7\times7$ view interpolation, with four corner views as input. We train and test our CILN for $2\times2\to7\times7$ view interpolation. This model is indicated as `Ours'  in Tab.~\ref{tab:2x2-7x7}.  In addition, we also compare with our CILN models trained for flexible spatial-angular super resolution (`Ours\textsuperscript{$\dagger$}') and variable pixel sparsity (`Ours\textsuperscript{$\dagger\dagger$}').  Since it is very challenging to recover EPI structures using only 4 corner views, the EPI-based approach of \cite{wuTIP2019} performs relatively poor, particularly in complex scenes containing occlusions and reflections. Fully convolutional approaches \cite{wing2018fast,meng2019high} and warping based approaches \cite{jin2019flexible, kalantari2016learning}, perform better indicating the advantage of higher dimensional convolutions and geometry based warping in effectively modeling the LF structure.
Our CILN trained for $2\times2\to7\times7$ view interpolation outperforms all the baselines, showing marked improvement in complex scenes containing occlusions and reflections, demonstrating the advantage of the proposed approach. While our CILN can be trained to recover LFs at flexible spatial resolutions (`Ours\textsuperscript{$\dagger$}' ), or from variable pixel sparsity `(Ours\textsuperscript{$\dagger\dagger$}'), this results in a slight degradation in performance on the fixed view interpolation task. 

Fig.~\ref{Fig:2x2_7x7} provides the visual comparison of the synthesized central view of the scene `Occlusion16'. Our approach can handle complex occlusions quite well as seen in the zoomed in patches. Error maps indicate superior reconstruction of our approach, which can preserve fine grained details. %Further, comparison with ground truth epi shows good preservation of LF parallax structure. 
    \begin{table}[htb]
        \centering
        \begin{tabular}{c c c c }
        \hline
        Method & 30scenes & Occlusions & Reflective\\
        \hline
        Meng \textit{et al.}\cite{meng2019high}& 39.25/0.970 & 35.72/0.964 & 35.62/0.945\\
        Yeung \textit{et al.}\cite{wing2018fast}&41.20/\textbf{0.982}&\textbf{36.92}/0.971&35.74/0.946\\
        %Kalantari \textit{et al.}\cite{kalantari2016learning}&37.50/0.970\\
        Ours\textsuperscript{*}&\textbf{41.30/0.982}&36.87/\textbf{0.972}&\textbf{36.02/0.949}\\
        \hline
        \end{tabular}
        \caption{Quantitative comparisons (PSNR/SSIM) of our proposed CILN with the state-of-the-art view synthesis approaches for the task $2\times2\rightarrow8\times8$ view interpolation. $*$ indicates our model was trained for $7\times7$ view interpolation.\label{tab:2x2-8x8}}
        \end{table}
\begin{figure}[htb]
 \begin{center}
 \resizebox{\linewidth}{!}{
\begin{tabular}{ll ll ll ll}
\multicolumn{2}{c}{\includegraphics[width=150pt]{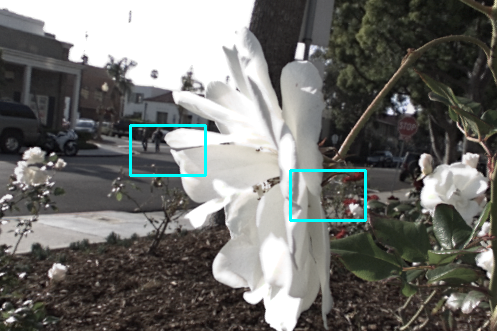}}&
\multicolumn{2}{c}{\hspace{-12pt}\includegraphics[width=150pt]{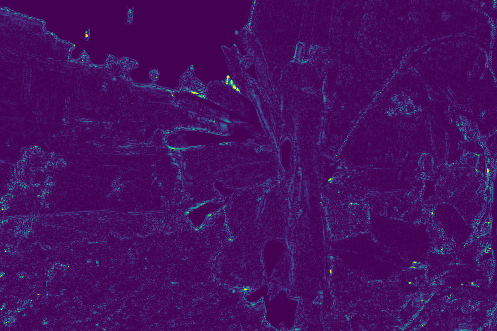}}&
\multicolumn{2}{c}{\hspace{-12pt}\includegraphics[width=150pt]{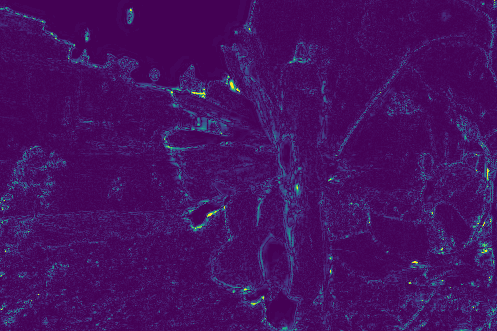}}&
\multicolumn{2}{c}{\hspace{-12pt}\includegraphics[width=150pt]{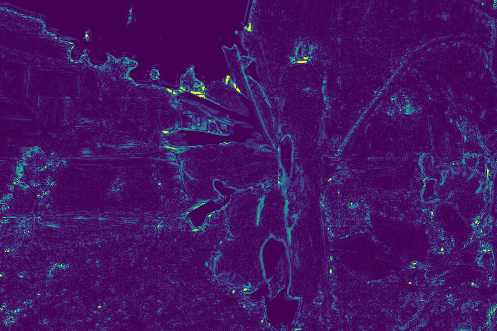}}\tabularnewline
\includegraphics[width=75pt]{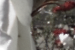}&\hspace{-12pt}
\includegraphics[width=75pt]{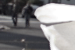}&\hspace{-12pt}
\includegraphics[width=75pt]{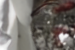}&\hspace{-12pt}
\includegraphics[width=75pt]{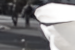}&\hspace{-12pt}
~\includegraphics[width=75pt]{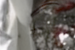}&\hspace{-12pt}
\includegraphics[width=75pt]{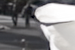}&\hspace{-12pt}
\includegraphics[width=75pt]{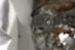}&\hspace{-12pt}
\includegraphics[width=75pt]{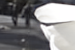}\\\\
\multicolumn{2}{c}{\huge Ground truth} &\multicolumn{2}{c}{\huge Ours} 
&\multicolumn{2}{c}{\huge  Yeung~etal.\cite{wing2018fast}}
&\multicolumn{2}{c}{\huge  Meng~etal.\cite{meng2019high}} 
\tabularnewline
\end{tabular}}
\caption{Comparison of synthesized views at view location $(4,4)$ of the LF `Flower2', for $2\times2\to8\times8$ view interpolation. Top row shows the comparison of error maps of CILN reconstruction and  recovered views using \cite{wing2018fast,meng2019high} with the ground truth view. Zoomed in patches corresponding to the marked regions are shown in the second row. Error maps are depicted with error magnified by a factor of $5$.}
\label{Fig:2x2_8x8}
\end{center}
\end{figure}
\\\textbf{{Flexible view interpolation:~}}To demonstrate the ability of CILN to recover LFs of flexible angular resolution, we also evaluate $2\times2\to8\times8$ LF recovery using the same model trained for $2\times2\to7\times7$ view interpolation. In Tab.~\ref{tab:2x2-8x8} and Fig.~\ref{Fig:2x2_8x8} we provide comparison with the fully convolutional approaches of Meng et~al. \cite{meng2019high} and Yeung et~al. \cite{wing2018fast} %and \blue{warping based approach of Kalantari et~al. \cite{kalantari2016learning}} 
using their publicly available code and trained checkpoints. Note that the baselines have an advantage as they are specifically trained for $8\times8$ view interpolation. Despite this, Tab.~\ref{tab:2x2-8x8} indicates that our CILN outperforms  the baselines in at least two test datasets, illustrating the benefit of the proposed approach. Visual comparison of error maps and zoomed in patches also indicates better reconstructions at the occlusion boundaries using our approach.
%\textbf{{View synthesis from irregular input views:~}}
    \begin{table}[htb]
        \centering
        \begin{tabular}{c c c c }
        \hline
        Method & $\times2$ & $\times3$ &$\times4$\\
        \hline
        Yeung \textit{et al.}\cite{wing2018fast}+Bicubic&35.98/0/947&32.91/0.895&31.08/0.849\\
        Yeung \textit{et al.}\cite{wing2018fast}+LIIF\cite{chen2020learning}&38.02/0.963&35.00/0.928&33.05/0.892\\
        Ours\textsuperscript{$\dagger$}&\textbf{38.69}/\textbf{0.968}&\textbf{35.90}/\textbf{0.940}&\textbf{33.99}/\textbf{0.908}\\
        \hline
        \end{tabular}
        \caption{Quantitative comparisons (PSNR/SSIM) of our proposed CILN with existing approaches for $2\times2\to8\times8$ angular and varying spatial upsampling on 30 scenes.\label{tab:spatial}}
        \end{table}
\subsection{Flexible spatial-angular super-resolution}
Feature resizing incorporated in our network architecture allows for reconstructions with flexible spatial resolutions. In our experiments, we train our CILN model ('Ours\textsuperscript{$\dagger$}') to recover  $7\times7$ views from variably downsampled $2\times2$ input views. We have seen in Tab.~\ref{tab:2x2-7x7} that this network performs view interpolation task, comparable to the baseline methods, while being slightly worse than our network trained without downsampling the inputs. To evaluate flexible spatio-angular super resolution, we test this CILN for the task of $2\times2\to8\times8$ view interpolation, with flexible spatial super-resolution factors without retraining. Since there are no other existing network-based baselines for flexible spatio-angular LF upsampling, we compare our scheme by sequentially applying angular and spatial super-resolution. That is, from the $2\times2$ input views, we first arrive at $8\times8$ LF views using the trained model of Yeung et~al. \cite{wing2018fast}. Subsequently, each of the $8\times8$ views are spatially upsampled using bicubic interpolation and the state-of-the-art flexible image super-resolution scheme of LIIF~\cite{chen2020learning}. Note that performing view interpolation before spatial upsampling preserves LF structure better than vice-versa. %While \cite{meng2019high} has attempted spatial-angular super resolution, it is only in a limited setting of $\times 2$ spatial and angular upsampling and therefore we do not compare with them.
\begin{figure*}[h]
 \begin{center}
\resizebox{\linewidth}{!}{
\begin{tabular}{ll ll ll ll ll ll ll}
 \multicolumn{2}{c}{\hspace{-12pt}\includegraphics[width=155pt]{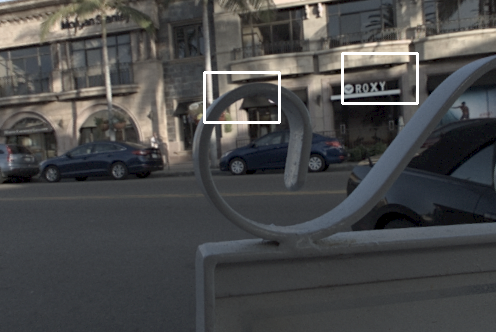}}&
\multicolumn{2}{c}{\hspace{-12pt}\includegraphics[width=155pt]{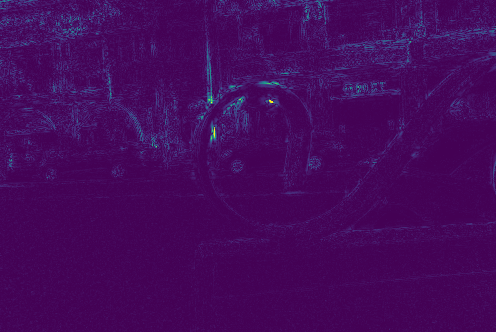}}&
\multicolumn{2}{c}{\hspace{-12pt}\includegraphics[width=155pt]{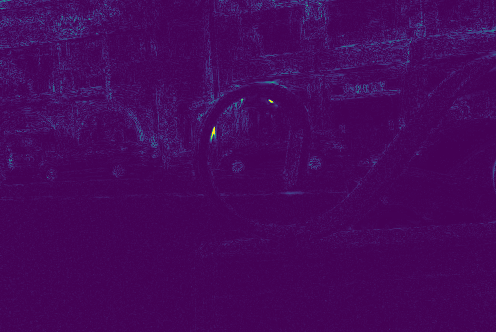}}&
\multicolumn{2}{c}{\hspace{-12pt}\includegraphics[width=155pt]{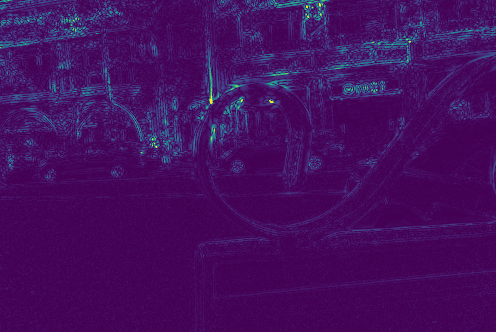}}&
\multicolumn{2}{c}{\hspace{-12pt}\includegraphics[width=155pt]{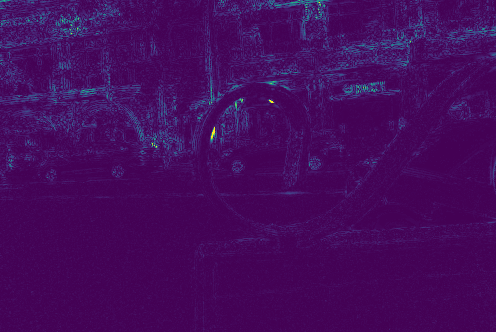}}&
\multicolumn{2}{c}{\hspace{-12pt}\includegraphics[width=155pt]{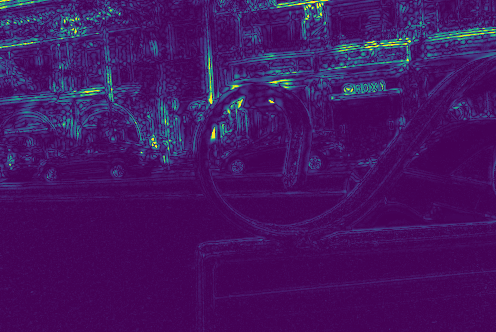}}&
\multicolumn{2}{c}{\hspace{-12pt}\includegraphics[width=155pt]{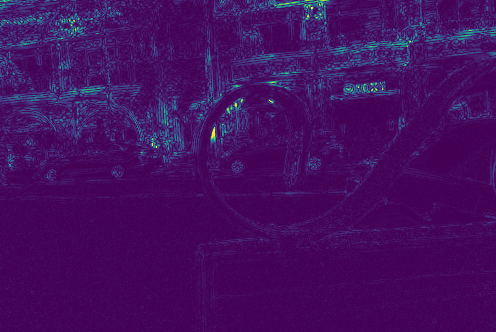}}\\
\hspace{-12pt}\includegraphics[width=74pt]{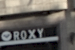}&\hspace{-11pt}
\includegraphics[width=74pt]{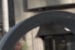}&\hspace{-11pt}
\includegraphics[width=74pt]{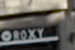}&\hspace{-11pt}
\includegraphics[width=74pt]{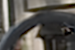}&\hspace{-11pt}
~\includegraphics[width=74pt]{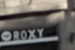}&\hspace{-11pt}
\includegraphics[width=74pt]{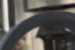}&\hspace{-11pt}
\includegraphics[width=74pt]{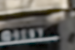}&\hspace{-11pt}
\includegraphics[width=74pt]{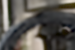}&\hspace{-11pt}
~\includegraphics[width=74pt]{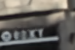}&\hspace{-11pt}
\includegraphics[width=74pt]{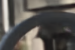}&\hspace{-11pt}
\includegraphics[width=74pt]{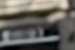}&\hspace{-11pt}
\includegraphics[width=74pt]{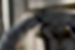}&\hspace{-11pt}
~\includegraphics[width=74pt]{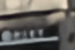}&\hspace{-11pt}
\includegraphics[width=74pt]{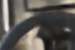}
\vspace{4pt}
\\
\multicolumn{2}{c}{ \Large Ground truth}
&\multicolumn{2}{c}{\hspace{-6pt}\Large Yeung et~al.~\cite{wing2018fast}+LIIF~\cite{chen2020learning}}
&\multicolumn{2}{c}{\Large CILN}
&\multicolumn{2}{c}{\hspace{-6pt}\Large Yeung et~al.~\cite{wing2018fast}+LIIF~\cite{chen2020learning}}
&\multicolumn{2}{c}{\Large CILN}
&\multicolumn{2}{c}{\hspace{-6pt}\Large Yeung et~al.~\cite{wing2018fast}+LIIF~\cite{chen2020learning}}
&\multicolumn{2}{c}{\Large CILN}\\
\multicolumn{2}{c}{}&
\multicolumn{4}{l}{\raisebox{.5\normalbaselineskip}[0pt][0pt]{$\underbrace{\phantom{\Large AAAAAAAAAAAAAAAAAAAAAAAAAAAAA}\hspace*{14\tabcolsep}}$}}&
\multicolumn{4}{l}{\raisebox{.5\normalbaselineskip}[0pt][0pt]{$\underbrace{\phantom{\Large AAAAAAAAAAAAAAAAAAAAAAAAAAAAA}\hspace*{14\tabcolsep}}$}}&
\multicolumn{4}{l}{\raisebox{.5\normalbaselineskip}[0pt][0pt]{$\underbrace{\phantom{\Large AAAAAAAAAAAAAAAAAAAAAAAAAAAAA}\hspace*{14\tabcolsep}}$}}\\
\multicolumn{2}{c}{}&\multicolumn{4}{c}{\Large$\times2$}&\multicolumn{4}{c}{\Large$\times3$}&\multicolumn{4}{c}{\Large$\times4$}
\end{tabular}}
\end{center}
\caption{Visual comparison of views at location $(4,4)$ for $\times4$ angular and $\times2$, $\times3$, $\times4$ spatial super-resolutions from correspondingly down sampled inputs for the scene `1586'. Shown are the ground truth view with marked regions indicating zoomed in patches, recovered views with our proposed CILN, and flexible spatial super-resolution using LIIF~\cite{chen2020learning} following view interpolation using Yeung~et al.~\cite{wing2018fast} .  Error maps are depicted with error magnified by a factor of 5.}
\label{Fig:spat_ang}
\end{figure*}
\begin{figure}[h]
 \begin{center}
 \begin{minipage}{0.21\linewidth}
 \centering
 \resizebox{\linewidth}{!}{
 \begin{tabular}{l}
{\hspace{-10pt}\rot{\large ~~~~~ p=50\% }~\includegraphics[width=90pt,height=62.5pt]{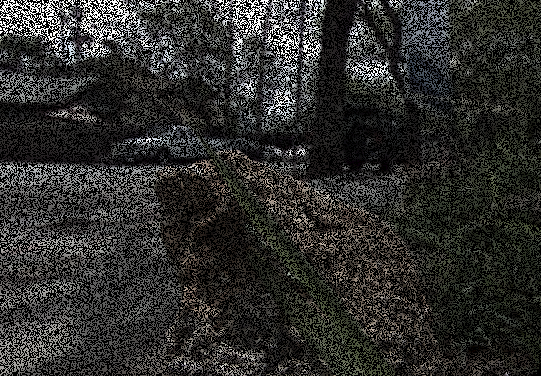}}\\
{\hspace{-10pt}\rot{\large ~~~~~ p=90\% }~\includegraphics[width=90pt,height=62.5pt]{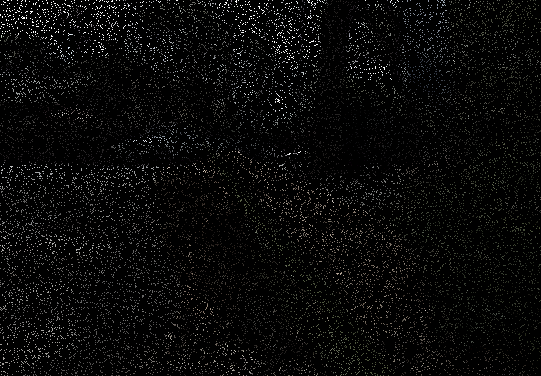}}\vspace{3pt}\\
\multicolumn{1}{c}{\large inputs}
\end{tabular}}
 \end{minipage}
\begin{minipage}{0.78\linewidth}
\resizebox{\linewidth}{!}{
\begin{tabular}{ll ll ll}
\multicolumn{2}{c}{\hspace{-18pt}\includegraphics[width=150pt]{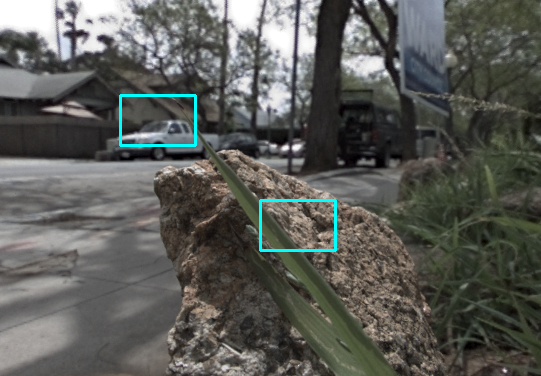}}&
\multicolumn{2}{c}{\hspace{-12pt}\includegraphics[width=150pt]{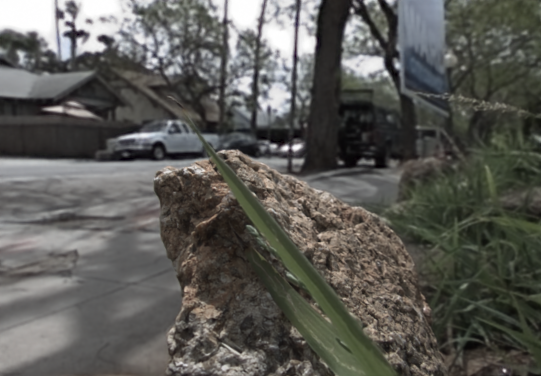}}&
\multicolumn{2}{c}{\hspace{-12pt}\includegraphics[width=150pt]{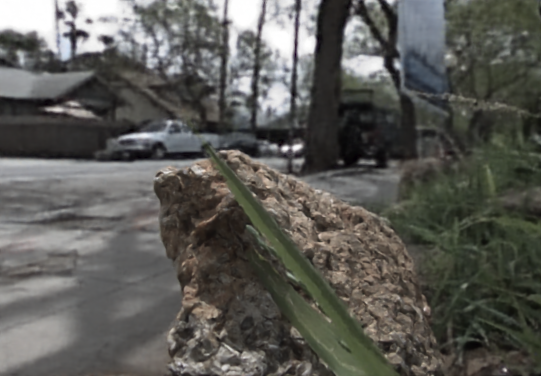}}\tabularnewline

\hspace{-15pt}\includegraphics[width=75pt]{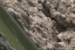}&\hspace{-12pt}
\includegraphics[width=75pt]{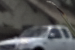}&\hspace{-12pt}
~\includegraphics[width=75pt]{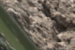}&\hspace{-12pt}
\includegraphics[width=75pt]{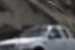}&\hspace{-12pt}
\includegraphics[width=75pt]{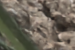}&\hspace{-12pt}
\includegraphics[width=75pt]{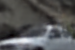}\vspace{3pt}\\
\multicolumn{2}{c}{\Large Ground truth} 
&\multicolumn{2}{c}{\Large Ours p=50\%}
&\multicolumn{2}{c}{\Large Ours p=90\%}
\tabularnewline
\end{tabular}}
\end{minipage}
\end{center}
\caption{Comparison of synthesized views at angular coordinates $(3,3)$ for the LF `Rock', with the ground truth view. First column shows an input view to our method with missing pixels. `p' indicates percentage of missing pixels. The columns 3 and 4 depict the reconstructed views using our CILN when $50\%$ and $90\%$ pixels are missing from input views. Zoomed in patches at the locations marked in the ground truth are shown.}
\label{Fig:pixel_drop}
\end{figure}

As seen in Tab.~\ref{tab:spatial}, our approach achieves the best performance, indicating the benefit of joint upsampling using implicit representation. Qualitative comparisons in Fig.~\ref{Fig:spat_ang} indicate lower error and better reconstruction of fine grain detail. In contrast, results with separate upsampling in angular and spatial domains using Yeung et~al. \cite{wing2018fast} and LIIF~\cite{chen2020learning} have significant artifacts at the occlusion boundaries. When both spatial and angular resolution of inputs are low, it becomes highly challenging to recover fine detail. In this case, even our approach also struggles to recover fine detail as observed in results of $\times4$ super-resolution in Fig.~\ref{Fig:spat_ang}.  
\subsection{Recovery from sparse spatio-angular inputs}
Another benefit of the proposed implicit network is its ability to recover LFs from measurements which are also spatially sparse. To evaluate this, we train our CILN model ('Ours\textsuperscript{$\dagger\dagger$}') on the task of $2\times2\to7\times7$ view interpolation, with $0-90\%$ of pixels, randomly dropped from input views. At test time,  when there is no pixel drop, this results in  only a marginal drop of performance compared to the CILN trained using clean input views (Tab.~\ref{tab:2x2-7x7}). The results of $2\times2\to7\times7$ view interpolation for varying extents of missing pixels in the input are reported in Tab.~\ref{tab:missing_small}. % Increasing pixel sparsity results in a gradual decay in reconstruction quality. 
Fig.~\ref{Fig:pixel_drop} depicts sample reconstruction of central view when input views have missing pixels. Even when 50\% pixels are missing from inputs, CILN demonstrates a fairly faithful recovery, capturing fine details and partial occlusions. The performance  degrades as expected when higher (90\%) pixels are missing. The zoomed in patch in Fig.~\ref{Fig:pixel_drop} shows the failure of CILN in recovering partial occlusion when 90\% pixels are missing from input views. Note, that image recovery from variable levels of missing pixels is a highly challenging task, and deep networks are generally trained separately for specific amounts of  degradation \cite{gao2017demand}. %Further, most view synthesis approaches cannot handle random missing pixels directly even with training.
\begin{table}[]
\centering
\begin{tabular}{lrrr}
\hline
p                         & 30scenes       & Occlusions     & Reflective \\ \hline
0.00   &42.34/0.985&38.83/0.979&38.89/0.960 \\
0.25 &   41.90/0.983&38.41/0.978 & 38.51/0.959\\
0.50& 41.02/0.980&37.54/0.973&37.86/0.954\\ 
0.75& 38.76/0.970 & 35.33/0.958&36.33/0.942\\ 
0.90 &34.59/0.934  &31.41/0.907  &33.23/0.908  \\ 
\hline
\end{tabular}
\caption{Average PSNR of novel views in dB for $7\times7$ view synthesis using CILN trained on varying number of missing pixels 0-90\%. p indicates the fraction of missing pixels.}
\label{tab:missing_small}
\end{table}
\begin{table}[!t]
        
        \centering
\resizebox{\linewidth}{!}{
\begin{tabular}{lllllll}
\hline
\multicolumn{2}{c}{Coordinate inputs}&MLP/&\multirow{2}{*}{Loss}&Flexible&\multirow{2}{*}{PSNR/SSIM}\\
$(s,t)$&($x,y)$&CNN&&output \\
\hline
\cmark&\cmark&MLP&L\textsubscript{1}+epi-loss&\cmark&42.80/0.986\\
\cmark&\xmark&MLP&L\textsubscript{1}+epi-loss&\cmark&41.27/0.983\\
\xmark&\xmark&CNN&L\textsubscript{1}+epi-loss&\xmark&42.37/0.984\\
\cmark&\cmark&MLP&L\textsubscript{1} loss&\cmark&42.52/0.985\\
\hline
\end{tabular}}
        \caption{Quantitative comparisons of CILN trained for $7\times7$ view interpolation  on 30 scenes test set with and without spatial coordinate inputs, with and without epi loss, and using implicit MLP decoder or CNN for view synthesis.}
        \label{tab:ablat_archi_train}
        \end{table}
       \begin{table}[!t]
       
        \centering
        \resizebox{\linewidth}{!}{
        \begin{tabular}{c c c c c }
        \hline
        Config.&Method &\hspace{-3pt} 30scenes & Occlusions & Reflective\\
        \hline
       \hspace{-10pt}\multirow{3}{*}{\parbox[c]{0pt}{
      \includegraphics[width=28pt]{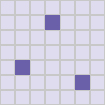}}}&\hspace{5pt}Kalantari \textit{et al.}\cite{kalantari2016learning}&40.86/0.981&36.63/0.970&38.77/0.954\\
        &Jin \textit{et al.}\cite{jin2019flexible}&42.57/0.986&39.12/0.980&40.00/0.961\\
        &Ours&43.70/0.987&41.01/0.984&41.52/0.968 \\\\
        \hspace{-10pt}\multirow{3}{*}{\parbox[c]{0pt}{
      \includegraphics[width=28pt]{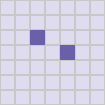}}}&Kalantari \textit{et al.}\cite{kalantari2016learning}&38.54/0.973&34.83/0.958&36.82/0.950\\
        &Jin \textit{et al.}\cite{jin2019flexible}&40.98/0.982&37.08/0.971&38.45/0.956\\
        &Ours&41.74/0.983&38.57/0.977&39.60/0.960 \\
        \hline
        \end{tabular}}
        \caption{Quantitative comparisons (PSNR/SSIM) of our approach with the view synthesis approaches \cite{kalantari2016learning,jin2019flexible} for $7\times7$ view synthesis from  input view configurations depicted in the first column. \label{tab:irregular}}
        \end{table}

\\\textbf{Ablation study:~}
In Tab.~\ref{tab:ablat_archi_train} we investigate the effectiveness of various components of our architecture and training for the task of $2\times2\to7\times7$ view interpolation. In our CILN formulation, we provided both  spatial coordinates $(x,y)$ and angular coordinates $(s,t)$ to the implicit network. We also evaluate the performance by CILN trained when only angular coordinates are input to CILN. We see that this results in a significant drop $(>1.5\text{dB})$ in PSNR, indicating the importance of providing the 4D coordinates. Further, we also replace the MLP implicit decoder with a two layer CNN having kernel size $1$ and $49\times3$ output channels corresponding to the three colors channels of the $7\times7$ views. Note that this does not have any coordinates as inputs and can only generate views on a fixed grid. Using such an architecture results in only a small drop in performance, showing the ability of simple 2D convolutional models in achieving competitive performance in fixed small baseline view interpolation. Our CILN model is trained using a combination of L\textsubscript{1} loss and EPI gradient loss~\cite{jin2020learning}. We also evaluate CILN trained using only L\textsubscript{1} loss between reconstruction and ground truth. As we can see in Tab.~\ref{tab:ablat_archi_train}, this results in a minor drop in performance compared to the original CILN trained using the combined loss.\vspace{1pt}\\
\textbf{Discussion: }
While we have mainly considered LF recovery from $2\times2$ input views, our approach can also handle irregular input views. Tab.~\ref{tab:irregular} shows the results of CILN for $7\times7$ LF recovery from irregular input views trained using the optimal sampling patterns from \cite{jin2019flexible} for tasks of $3\to7\times7$ and $2\to7\times7$ LF recovery.  We compare with the warping based approaches \cite{jin2019flexible}, \cite{kalantari2016learning} since they  can handle  irregular input view configurations. As we can see in Tab.~\ref{tab:irregular}, our approach can also recover high quality LFs from  as low as only 2 input views which are irregularly sampled. Note that models of \cite{jin2019flexible}, \cite{kalantari2016learning} are trained for LF recovery from flexible configuration of fixed number of views. Our training with fixed pattern sampling likely gives our CILN an advantage.  We will further investigate how to empower the CILN approach to handle flexible sampling patterns  in future work.
\begin{table}[]
\centering
\begin{tabular}{r|l|l}
Algorithm         & Running Time & GPU Memory\\
\hline
Meng~et~al.\cite{meng2019high} & 620 ms $\pm$ 5.37 ms & 5776 MiB\\ 
Jin~et~al.\cite{jin2019flexible} & 7.52 s $\pm$ 14.4 ms & 8788 MiB\\ 
Ours & 237.5 ms $\pm$ 9.32 ms & 10602 MiB 
\end{tabular}
\caption{Mean running time $\pm$ std. dev. for  $2\times 2$ to $7\times 7$  view interpolation over 10 runs. Timing and memory consumption of \cite{meng2019high,jin2019flexible} are reported for only single channel (Y channel) view synthesis, whereas the numbers reported  are for view synthesis in all three RGB channels for our approach. }
\label{tab:run_time}
\end{table}
\vspace{2pt}\\\textbf{Running time:}
We compare the running time and memory consumptions of our method with the methods of Jin~et~al. \cite{jin2019flexible} and Meng~et~al. \cite{meng2019high} for the task of $2\times 2$ to $7\times 7$ view recovery. This comparison was done on a machine with an Intel i9-7900X CPU @ 3.30 GHz, 128 GB RAM and an NVIDIA RTX 2080Ti GPU. For the run time experiments only, we used input LFs with a patch size of $2\times 2\times 200\times 200$. The results in Tab.~\ref{tab:run_time} show that the simple 2D convolutional architecture of our network allows a much lower computation time compared to the other two methods. % albeit with a higher memory consumption compared to Meng~et~al.\cite{meng2019high}. 
Note that, our model outputs all the three color channels in the RGB space while the outputs of the \cite{jin2019flexible} and \cite{meng2019high} correspond to only the luminence component in the YCbCr space.
%, which is the largest size that could fit in our GPU memory for all tested methods
%-------------------------
\section{Conclusions}
In this paper, we  presented  conditional implicit light field networks, a novel deep implicit  representation  for LFs, which generalizes across scenes. 
%Given sparse spatio-angular observations on a regular grid, 
Given sparse input views, our CILN  predicts the scene radiance at any queried point in the spatial and angular dimensions. Our framework  achieves this by propagating the input context through a convolutional neural network, to provide  pixel level local fused embeddings. Our implicit network exploits these local embeddings to capture fine-grained details and generate a photorealistic LF reconstruction. Qualitative and quantitative experiments validate that our CILN can provide reconstructions outperforming recent state of the art approaches for LF view synthesis on real scenes. Our CILN can generate LF views at arbitrary  spatio-angular resolutions clearly demonstrating our flexibility. Further, CILN can also be trained to be robust to varying levels of spatial sparsity with a single model. Future work may include extending such implicit representations to much larger baseline light fields.

\ifCLASSOPTIONcaptionsoff
  \newpage
\fi

\end{document}